\documentclass[11pt]{article}

\usepackage[final]{acl}

\usepackage{times}
\usepackage{latexsym}

\usepackage[T1]{fontenc}

\usepackage[utf8]{inputenc}

\usepackage{microtype}

\usepackage{inconsolata}

\usepackage{graphicx}
\usepackage{comment}

\usepackage{lipsum}
\usepackage{enumitem}
\hypersetup{
  colorlinks   = true, 
  urlcolor     = purple, 
  linkcolor    = magenta, 
  citecolor   = purple 
}
\usepackage{booktabs}
\usepackage{makecell}
\usepackage{hyperref}
\usepackage{float}   
\usepackage{subcaption} 
\usepackage{longtable}
\usepackage{array}
\usepackage{enumitem}
\usepackage{multicol}
\usepackage{listings}
\usepackage{minted}
\usepackage{fvextra}
\DefineVerbatimEnvironment{wrapverbatim}{Verbatim}{breaklines=true}

\usepackage{tabularx}

\title{Heterogeneity in Formal Linguistic Competence\\ of Language Models: Is Data the Real Bottleneck?}

\author{H S V N S Kowndinya Renduchintala \\
  Media and Data Science Research \\
  Adobe Inc., India \\
  \texttt{kowndinya.renduchintala@gmail.com} \\\And
  Sumit Bhatia \\
  Media and Data Science Research \\
  Adobe Inc., India \\
  \texttt{sumit.bhatia@adobe.com} \\}

\begin{document}
\maketitle
\begin{abstract}
Large Language Models (LLMs) exhibit a puzzling disparity in their formal linguistic competence: while they learn some linguistic phenomena with near-perfect mastery, they often perform below chance on others, even after training on trillions of tokens. In this work, we investigate whether these failures stem from inherent architectural limitations or simply the scarcity of these specific grammatical constructions in web-scale corpora. We pre-train simple GPT-2 Small (124M) models on a 100M-token random sample of the FineWeb corpus and intervene by injecting a minimal amount (1\%) of synthetic data targeting specific linguistic phenomena. We find that this targeted intervention substantially improves model performance in 8 out of the 9 worst-performing BLiMP paradigms -- notably the accuracy on a specific paradigm, only\_npi\_scope, surges from 20.9\% to 69.4\%. Furthermore, we observe that these interventions generally preserve or slightly improve aggregate performance. However, while we also identify a resistant phenomenon, principle\_A\_c\_command, whose performance remains below chance even after our data augmentation, our findings do serve as an optimistic existence proof that even small language models can substantially improve on those linguistic phenomena on which models typically perform poorly, provided the pre-training data contains sufficient exposure to them. This suggests that efforts towards human-scale language modeling may benefit greatly by focusing on data composition. The code to reproduce our results is open-sourced at \href{https://github.com/kowndinya-renduchintala/heterogeneity-in-formal-linguistic-competence}{https://github.com/kowndinya-renduchintala/heterogeneity-in-formal-linguistic-competence}.
\end{abstract}

\section{Introduction}
\label{sec:introduction}

Large Language Models (LLMs), pre-trained on web-scale text corpora with a simple next-token prediction objective~\citep{olmo2025olmo,achiam2023gpt,brown2020language}, now underpin most state-of-the-art language technologies. The most notable among their remarkable capabilities, however, is their unprecedented fluency in generating human-like natural language; in many contexts, their generated text is now indistinguishable from that of humans~\citep{jones2025people}.

This linguistic mastery positions them as a critical empirical testing ground for the Poverty of Stimulus arguments~\citep{lappin2007machine}, by serving as an existence proof that linguistic knowledge \textit{can} emerge solely from massive statistical learning, without any innate linguistic priors. Furthermore, recent evidence~\citep{hu2024findings} has shown that complex grammar can be learned even from as low as 100M pre-training tokens -- amount of data similar to what a 13-year-old child encounters in their entire life~\citep{gilkerson2017mapping}.

\begin{table*}[ht!]
\centering
\resizebox{\textwidth}{!}{%
\begin{tabular}{llc|cccc}
\toprule
Model & \makecell{Pre-Training \\ Data} & \makecell{BLiMP \\ (Aggregate)} & \makecell{only\\ npi\_scope} & \makecell{principle\_A\\ reconstruction} & \makecell{existential\_there\\quantifiers\_2} & \makecell{sentential\\subject\\island} \\
\midrule
GPT-2 124M~\citep{radford2019language} & FineWeb-100M & 74.77 & 20.9 & 37.2 & 23.4 & 40.4 \\
GPT-2 124M~\citep{radford2019language} & FineWeb-10B  & 80.13 & 57.8 & 36.2 & 23.2 & 44.5 \\
SmolLM2 135M~\citep{allal2025smollm2} & FineWeb-15T  & 80 & 44 & 33.7 & 33.5 & 45.4 \\
SmolLM2 1.7B~\citep{allal2025smollm2} & FineWeb-15T & 81.51 & 63.2 & 53.6 & 38.4 & 42.1 \\
Llama-3 8B~\citep{grattafiori2024llama} & 15T+ & 83.39 & 61.3 & 55 & 62.3 & 56.4 \\ 
Llama-3 70B~\citep{grattafiori2024llama} & 15T+ & 84.16  & 63.8  & 49.6 & 57.6 & 49.9 \\
\bottomrule
\end{tabular}
}
\caption{Contrast between aggregate scores and performance on some specific BLiMP paradigms that models typically perform poorly on.}
\label{tab:blimp_results}
\end{table*}

Various benchmarks have been proposed to evaluate the extent to which a language model has acquired \textit{formal linguistic competence}~\citep{mahowald2023dissociating} -- i.e., the knowledge of rules and statistical regularities of language  -- among which, BLiMP~\citep{warstadt2020blimp} is a linguistically motivated benchmark that probes this knowledge across a wide range of linguistic phenomena. This suite isolates specific syntactic paradigms by testing whether a model assigns higher probability to a grammatical sentence than to its minimally different ungrammatical counterpart.

While many contemporary language models (including those pre-trained on human-scale datasets) achieve impressive aggregate scores on this benchmark, taking a closer look exposes stark disparities at the level of individual linguistic phenomena, with models demonstrating near-perfect mastery of some linguistic phenomena, while performing significantly below chance on others. More critically, these failures do not get solved by simply scaling the number of pre-training tokens. For specific paradigms such as \texttt{principle\_A\_reconstruction}, we even observe an inverse U-shaped scaling curve, where performance improves during early training stages but paradoxically deteriorates as the model consumes more tokens (Appendix~\ref{app:learning_curves}).

A plausible reason for this heterogeneity in the formal linguistic competence of LMs could be the relatively low representation of some of these specific linguistic phenomena in typical web corpora. While it has been shown that a rare phenomenon can still be learned from related, less rare constructions~\citep{misra2024language}, it is still not clear if poor performance on a specific paradigm can be attributed to lack of data alone.

Since precisely estimating the frequency of occurrence of a specific linguistic phenomenon in text corpora is challenging, in this paper we systematically test this by studying the effect of controlled data interventions that inject a small fraction ($1\%$) of natural-looking synthetic text containing sentences that instantiate a target BLiMP paradigm, into the pre-training data. If data, rather than other factors such as model architecture, is the primary bottleneck for learning the corresponding grammatical rule, we would expect to observe a significant improvement in performance on the target paradigm.

To summarize our contributions, in this work,
\begin{itemize}[nosep]
    \item We identify 9 failure cases of BLiMP paradigms where a standard GPT-2 model performs below chance and introduce a targeted intervention methodology using a minimal injection (1\%) of natural-looking, syntactically-controlled synthetic text during pre-training.
    \item We demonstrate that data scarcity is the primary bottleneck for 8 out of 9 paradigms, with our intervention yielding substantial accuracy gains (up to +48.5 points absolute gain) while generally preserving or even slightly improving the aggregate benchmark performance.
\end{itemize}

\newpage
\section{Background}
\label{sec:background}
While many contemporary language models achieve strong aggregate BLiMP scores, their competence is highly uneven at the level of individual linguistic paradigms. Table~\ref{tab:blimp_results} illustrates this heterogeneity: models with impressive aggregate performance often exhibit dramatic failures on specific paradigms, such as only\_npi\_scope, principle\_A\_reconstruction, existential\_there\_quantifiers\_2 and sentential\_subject\_island, among many others. Importantly, these weaknesses are not reliably resolved by increasing the amount of pre-training data, with some paradigms showing little improvement -- or even degradation -- despite increasing the number of pre-training tokens by orders of magnitude.

The learning curves in Appendix~\ref{app:learning_curves} demonstrate that different linguistic phenomena are learned at markedly different rates when a GPT-2 (124M) model is pre-trained on up to 10B tokens randomly sampled from the FineWeb corpus~\citep{penedo2024fineweb}, and that some paradigms fail to be learned altogether. As noted in Section~\ref{sec:introduction}, for specific paradigms such as principle\_A\_reconstruction, we even observe
an inverse U-shaped scaling curve. Whether these failure cases can be explained by data scarcity -- i.e., insufficient exposure to the relevant constructions during pre-training -- remains unclear.

\section{Approach}
Since our goal is to test whether failures in formal linguistic competence arise from insufficient exposure to particular grammatical constructions or from more fundamental representational limitations of the model architecture, we design a minimal, controlled data intervention that selectively increases exposure to a single linguistic paradigm while leaving all other aspects of pre-training such as model architecture, training hyperparameters, total number of pre-training tokens etc., unchanged. More specifically, we use an LLM to generate a tiny synthetic corpus comprising a few documents -- each containing a sentence that demonstrates the given linguistic paradigm (as defined by its syntactic template). Then, we inject a small fraction ($\sim 1\%$) of this data into our pre-training dataset.

To maintain diversity in the generated documents, we curate a taxonomy of genres (and subgenres) and a list of lemmas; and while generating each document, we randomly select a genre (and a subgenre) and a few lemmas from these lists, and instruct the LLM to generate text that reads like a genuine piece of writing in the specified genre (and subgenre) while also ensuring that the given set of lemmas all appear in the generated text.

A significant performance gain on the target paradigm following this intervention would serve as an existence proof that the model possesses the necessary inductive biases to learn the construction, identifying data scarcity -- rather than architectural constraints, etc. -- as a more significant bottleneck.

\section{Experimental Setup}
\label{sec:experimental_setup}
\begin{table*}[ht!]
\centering
\small
\begin{tabular}{lcccc}
\toprule
Linguistic Paradigm & Baseline & Random & Targeted & $\Delta$ \\
\midrule
\texttt{only\_npi\_scope} & 20.9 & 17.0 & \textbf{69.4} & +48.5 \\
\texttt{existential\_there\_quantifiers\_2} & 23.4 & 12.6 & \textbf{51.8} & +28.4 \\
\texttt{wh\_vs\_that\_with\_gap\_long\_distance} & 23.8 & 24.6 & \textbf{41.2} & +17.4 \\
\texttt{matrix\_question\_npi\_licensor\_present} & 30.8 & 25.2 & \textbf{44.2} & +13.4 \\
\texttt{principle\_A\_reconstruction} & 37.2 & 44.7 & \textbf{78.9} & +41.7 \\
\texttt{sentential\_subject\_island} & 40.4 & 31.3 & \textbf{52.2} & +11.8 \\
\texttt{left\_branch\_island\_echo\_question} & 45.3 & 44.6 & \textbf{63.2} & +17.9 \\
\texttt{coordinate\_structure\_constraint\_complex\_left\_branch} & 46.5 & 40.7 & \textbf{62.4} & +15.9 \\
\texttt{principle\_A\_c\_command} & 49.0 & 50.4 & 46.0 & -3.0 \\
\bottomrule
\end{tabular}
\caption{Accuracy on the 9 target paradigms on which Baseline has the worst (below chance) performance. Comparison between Baseline (100M FineWeb), Random (1\% random synthetic), and Targeted (1\% specific synthetic). $\Delta$ shows the absolute improvement of Targeted over Baseline.}
\label{tab:intervention_results}
\end{table*}
\subsection{Synthetic Data Generation}
For synthetic data generation, we use GPT-OSS 120B~\citep{agarwal2025gpt} as our underlying LLM -- we set its reasoning effort to medium, and to encourage diversity in the generated text, we set the temperature to $0.7$. The stylistic diversity of the generated documents is controlled via randomly sampling from a curated taxonomy of genres and subgenres (Appendix~\ref{app:synthetic_data_generation_genres}), while lexical diversity is ensured by sampling from a pool of the $10,000$ most frequent non-stopword lemmas extracted from the Google Web Trillion Word Corpus~\citep{norvig_ngrams}. The prompt used for synthetic data generation is provided in Appendix~\ref{app:synthetic_data_generation_prompt} and the syntactic templates corresponding to the linguistic paradigms we consider are provided in Appendix~\ref{app:synthetic_data_generation_templates}.

\subsection{Base Model and Pre-Training}

For all our experiments, we use GPT-2 Small 124M~\citep{radford2019language}, a simple autoregressive model architecture with FlashAttention-2~\citep{dao2023flashattention} and a maximum sequence length of 1024. Models are pre-trained on a total budget of 100M tokens for 20 epochs, with an effective batch size of 128, a learning rate of $6 \times 10^{-4}$ with cosine decay, a warmup fraction of 0.05, and a weight decay of 0.01. The baseline model is pre-trained on 100M tokens sampled exclusively from the FineWeb corpus~\citep{penedo2024fineweb}. For our intervention studies, we replace 1\% of the baseline corpus with the generated synthetic data, resulting in a training set of 99M FineWeb tokens and 1M synthetic tokens. All hyperparameters remain constant across runs to isolate the effect of the data intervention. Also, note that out of the 1M synthetic tokens, the number of sentences that actually demonstrate a target BLiMP paradigm is only around the order of $\sim 3$K. This is because we prompted the LLM to generate full, naturalistic documents in various genres and the target construction only appears embedded within a longer document. 

\subsection{Evaluation Protocol}
We evaluate all the models on the full BLiMP suite~\citep{warstadt2020blimp}, which is a benchmark for evaluating the grammatical knowledge of language models. It consists of minimally different pairs of sentences -- one grammatical and one ungrammatical -- designed to isolate specific linguistic phenomena. Models are evaluated based on whether they assign higher probability to the grammatical sentence in each pair. It consists of 67 individual paradigms spanning 12 linguistic phenomena in English.

\section{Results}
\label{sec:results_n_discussion}

We evaluate the effect of targeted data interventions on the 9 worst-performing BLiMP paradigms (that correspond to below-chance accuracies) for a GPT-2 Small model trained on 100M tokens of FineWeb. We compare the performance of our targeted intervention models against two baselines: the base model trained on 100M FineWeb tokens (Baseline) and the control model trained with 1\% random synthetic data (Random).

\subsection{Targeted Data Interventions: Impact on Paradigm-Specific Performance}

As summarized in Table~\ref{tab:intervention_results}, targeted data interventions led to substantial improvements in 8 out of the 9 worst-performing BLiMP paradigms that we considered. The most dramatic gains were observed in \texttt{only\_npi\_scope}, where accuracy surged from 20.9\% (baseline) to 69.4\% (+48.5 points), and \texttt{principle\_A\_reconstruction}, which improved from 37.2\% to 78.9\% (+41.7 points). These results strongly support the hypothesis that the model’s initial failures are not due to inherent architectural limitations, but rather to insufficient exposure to the relevant grammatical constructions during pre-training with naturally occurring corpora.
For most of the paradigms, the intervention effectively brought the models up from "failure modes" (accuracies of below 50\%) to performance levels significantly above chance. Moreover, even though performance on a couple of paradigms such as \texttt{wh\_vs\_that\_with\_gap\_long\_distance} and \texttt{matrix\_question\_npi\_licensor\_present} still remains below-chance after targeted intervention, they are nevertheless substantially improved compared to the Baseline, showing that exposure to more data that demonstrate these paradigms helps in learning them better.

The only paradigm that did not benefit from intervention was \texttt{principle\_A\_c\_command}. Performance remained below-chance levels (Baseline: 49.0\%, Targeted: 46.0\%). Unlike \texttt{principle\_A\_reconstruction}, which saw substantial gains, this paradigm was more resistant to our minimal data injection. While the reasons are unclear, we hypothesize that this paradigm may be particularly challenging for the model due to the long-range dependency combined with presence of a linearly very close distractor noun; and might require presence of significantly more such sentences in training data.

\paragraph{Comparison with Random Baseline} The Random Baseline tracked closely with the FineWeb baseline across all tasks, and in some cases (e.g., \texttt{existential\_there\_quantifiers\_2}, 12.6\%) even underperformed it. This confirms that the improvements observed in the targeted models are not due to the mere presence of synthetic text or a change in the token distribution, but are directly attributable to the specific syntactic evidence provided by the targeted examples. Moreover, all these gains are achieved with an extremely small data intervention -- only 1\% of the total training data ($1$M tokens consisting of $\sim 3$K sentences of the specific paradigm).

\subsection{Targeted Data Interventions: Impact on Aggregate Performance}

Beyond the paradigm-specific improvements, we examine the impact of these data interventions on the models' broader linguistic competence since overrepresenting specific syntactic templates could lead to catastrophic forgetting of other linguistic phenomena or degradation in general grammatical proficiency.

Contrary to such concerns, we observe that targeted interventions generally preserve or slightly improve global model performance (refer to Appendix~\ref{app:detailed_results} for a detailed breakdown of results), suggesting that the injection of 1\% syntactically diverse synthetic data is sufficiently minimal to avoid distorting the model’s general distribution while being effective enough to fix specific deficiencies in linguistic knowledge.


\subsection{Effect of Varying the Intervention Magnitude}
\label{sec:ablation_study}
\begin{table*}[t]
    \centering
    \small
    \begin{tabular}{lrrrr}
    \toprule
    \textbf{Paradigm} & \textbf{Baseline} & \textbf{0.01\%} & \textbf{0.1\%} & \textbf{1\%} \\
    \midrule
    \texttt{only\_npi\_scope} & 20.9 & 41.9 & 31.0 & \textbf{69.4} \\
    \texttt{existential\_there\_quantifiers\_2} & 23.4 & 13.2 & 25.2 & \textbf{51.8} \\
    \texttt{wh\_vs\_that\_with\_gap\_long\_distance} & 23.8 & 26.4 & 28.6 & \textbf{41.2} \\
    \texttt{matrix\_question\_npi\_licensor\_present} & 30.8 & 38.1 & 35.1 & \textbf{44.2} \\
    \texttt{principle\_A\_reconstruction} & 37.2 & 72.3 & 71.3 & \textbf{78.9} \\
    \texttt{sentential\_subject\_island} & 40.4 & 40.5 & 43.9 & \textbf{52.2} \\
    \texttt{left\_branch\_island\_echo\_question} & 45.3 & 49.3 & 52.5 & \textbf{63.2} \\
    \texttt{coordinate\_structure\_constraint\_complex\_left\_branch} & 46.5 & 52.7 & 59.3 & \textbf{62.4} \\
    \texttt{principle\_A\_c\_command} & 49.0 & 46.1 & 49.6 & 46.0 \\
    \bottomrule
    \end{tabular}
    \caption{Effect of varying the fraction of targeted synthetic data injected during pre-training. Bold indicates the best accuracy for each paradigm.}
    \label{tab:ablation_results}
\end{table*}

To better understand how the amount of targeted synthetic data affects performance, we conduct an ablation study where we vary the fraction of injected synthetic data across three levels: 0.01\%, 0.1\%, and 1\% of the total pre-training tokens. Table~\ref{tab:ablation_results} summarizes the results.

For most paradigms, we observe a consistent trend of improving performance as the fraction of synthetic data increases. \texttt{coordinate\_structure\_constraint\_complex\_}\\\texttt{left\_branch}, for instance, improves steadily from 46.5\% (baseline) to 52.7\%, 59.3\%, and 62.4\% across the three intervention levels. A particularly striking case is \texttt{principle\_A\_reconstruction}, where even a 0.01\% injection (corresponding to roughly 10K synthetic tokens containing ~30 examples of the paradigm) is sufficient to boost accuracy from 37.2\% to 72.3\%, suggesting that the model requires remarkably little evidence to acquire this construction. However, not all paradigms respond monotonically -- such as \texttt{only\_npi\_scope} and \texttt{existential\_there\_quantifiers\_2}. These anomalies suggest that the interaction between targeted synthetic data and existing corpus statistics may be more complex for certain paradigms. However, \texttt{principle\_A\_c\_command} remains at below-chance levels across all intervention magnitudes.

\section{Discussion}
\label{sec:discussion}
Our results carry implications beyond the immediate question of improving benchmark performance. Language models have long been used as models of human language processing. Yet, a persistent criticism is that their data requirements -- often trillions of tokens -- are unrealistically large compared to the input available to human learners~\citep{gilkerson2017mapping}. Our finding that a 124M-parameter model pre-trained on just 100M tokens with tiny data intervention can outperform a 70B-parameter model trained on over 15 trillion tokens on specific paradigms (e.g., \texttt{principle\_A\_reconstruction}: 78.9\% vs. 49.6\% for Llama-3 70B) suggests that data composition may matter more than data scale for acquiring formal linguistic competence. We also observe that targeting a single paradigm often improves related, non-targeted paradigms (see Appendix~\ref{app:detailed_results}). These findings lend support to efforts like the BabyLM challenge~\citep{hu2024findings} that explore pre-training at human-scale data budgets and indicate that small models can acquire robust grammatical knowledge provided the training data contains sufficient distributional evidence for the relevant constructions.

\section{Conclusion}
We demonstrated that the heterogeneity in the formal linguistic competence of LLMs largely stems from data scarcity rather than inherent architectural limitations etc. By injecting a minimal (1\%) signal of targeted synthetic data into the pre-training of GPT-2, we substantially improved performance on 8 out of 9 previously failing paradigms -- such as principle\_A\_reconstruction and only\_npi\_scope -- and also observed that these targeted interventions generally preserve or slightly improve aggregate model performance. However, the persistent failure of principle\_A\_c\_command suggests that some structural constraints may require significantly higher data density or stronger inductive biases to acquire. Ultimately, our findings serve as an optimistic existence proof: standard language models possess the latent capacity to acquire formal syntax, provided the pre-training data offers sufficient signal, suggesting that efforts towards human-scale language modeling may benefit greatly by focusing on getting the data composition right.

\section*{Limitations}

Our experiments rely on GPT-2 Small (124M parameters) trained on a relatively modest 100M-token budget, which differs substantially from modern LLMs trained on trillions of tokens. Although we observe that increasing data scale alone does not resolve certain failures in larger models such as Llama-3 (Table~\ref{tab:blimp_results}), we do not test whether our intervention strategy scales similarly to billion-parameter models. Moreover, testing on a single model architecture limits the conclusions we can draw about architectural limitations -- changing the parameter count or the architecture itself could affect how well a model learns from sparse signals, and we leave this exploration to future work.

Second, while our ablation study (Section~\ref{sec:ablation_study}) explores intervention magnitudes of 0.01\%, 0.1\%, and 1\%, we do not explore the upper range -- it remains unclear whether larger fractions of targeted data could resolve persistent failures such as \texttt{principle\_A\_c\_command}. Additionally, we conduct targeted interventions for only 9 of the 67 BLiMP paradigms; extending this would provide a more complete picture. Finally, we observe that targeting one paradigm can both help and hurt performance on other paradigms (Appendix~\ref{app:detailed_results}), but understanding why certain grammatical phenomena transfer positively or negatively would require deeper investigation into the model's internal representations -- a direction we leave for future work.


\bibliography{references}

\onecolumn
\appendix

\section{Code \& Data}
Our code that we used for pre-training and for generating the synthetic data for data intervention experiments is open-sourced at \href{https://github.com/kowndinya-renduchintala/heterogeneity-in-formal-linguistic-competence}{https://github.com/kowndinya-renduchintala/heterogeneity-in-formal-linguistic-competence}. 

\section{Learning Curves of BLiMP paradigms}
\label{app:learning_curves}
\begin{figure}[H]
\centering

\begin{minipage}[t]{0.48\linewidth}
    \centering
    \includegraphics[width=\linewidth]{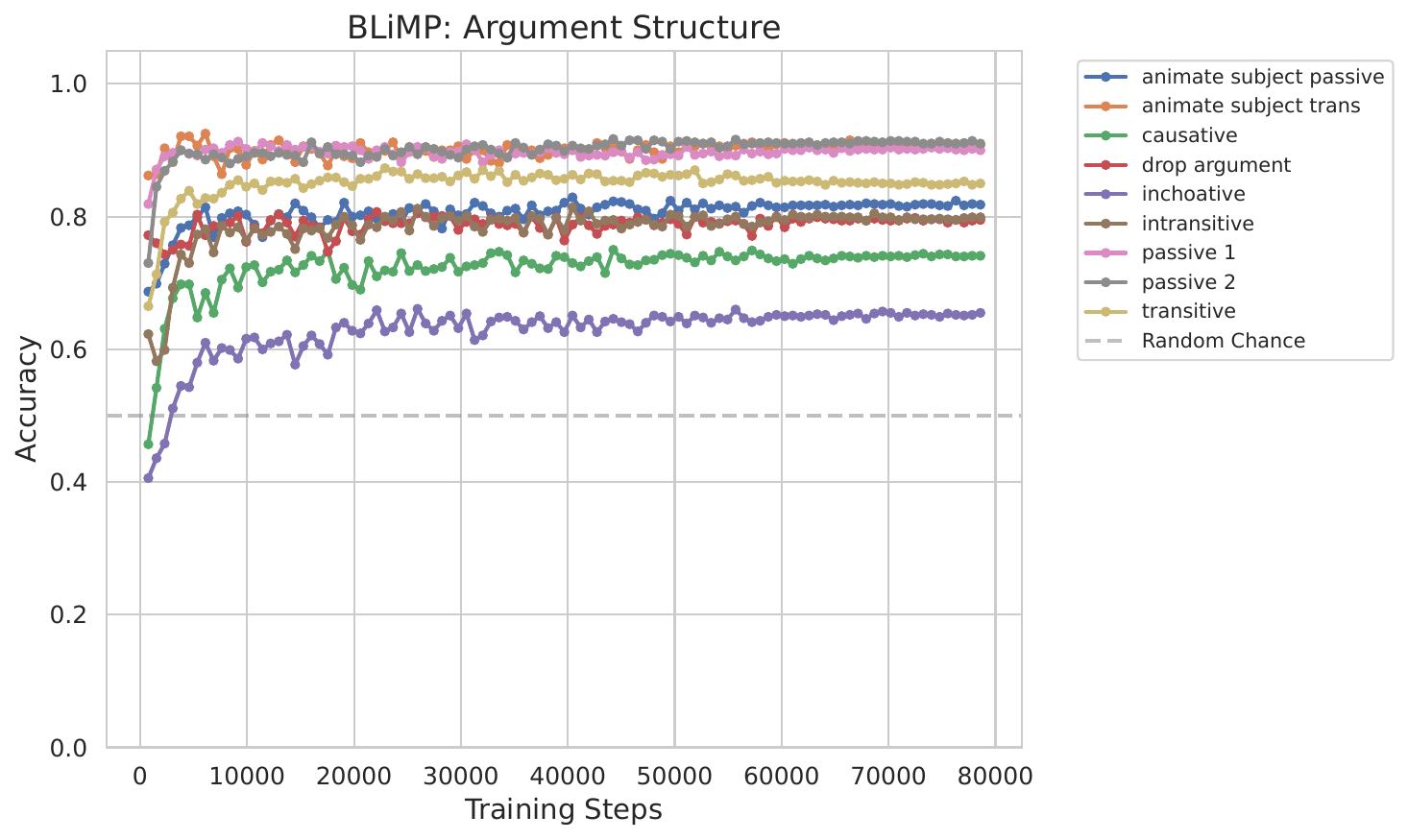}
    \caption*{Argument Structure}
\end{minipage}\hfill
\begin{minipage}[t]{0.48\linewidth}
    \centering
    \includegraphics[width=\linewidth]{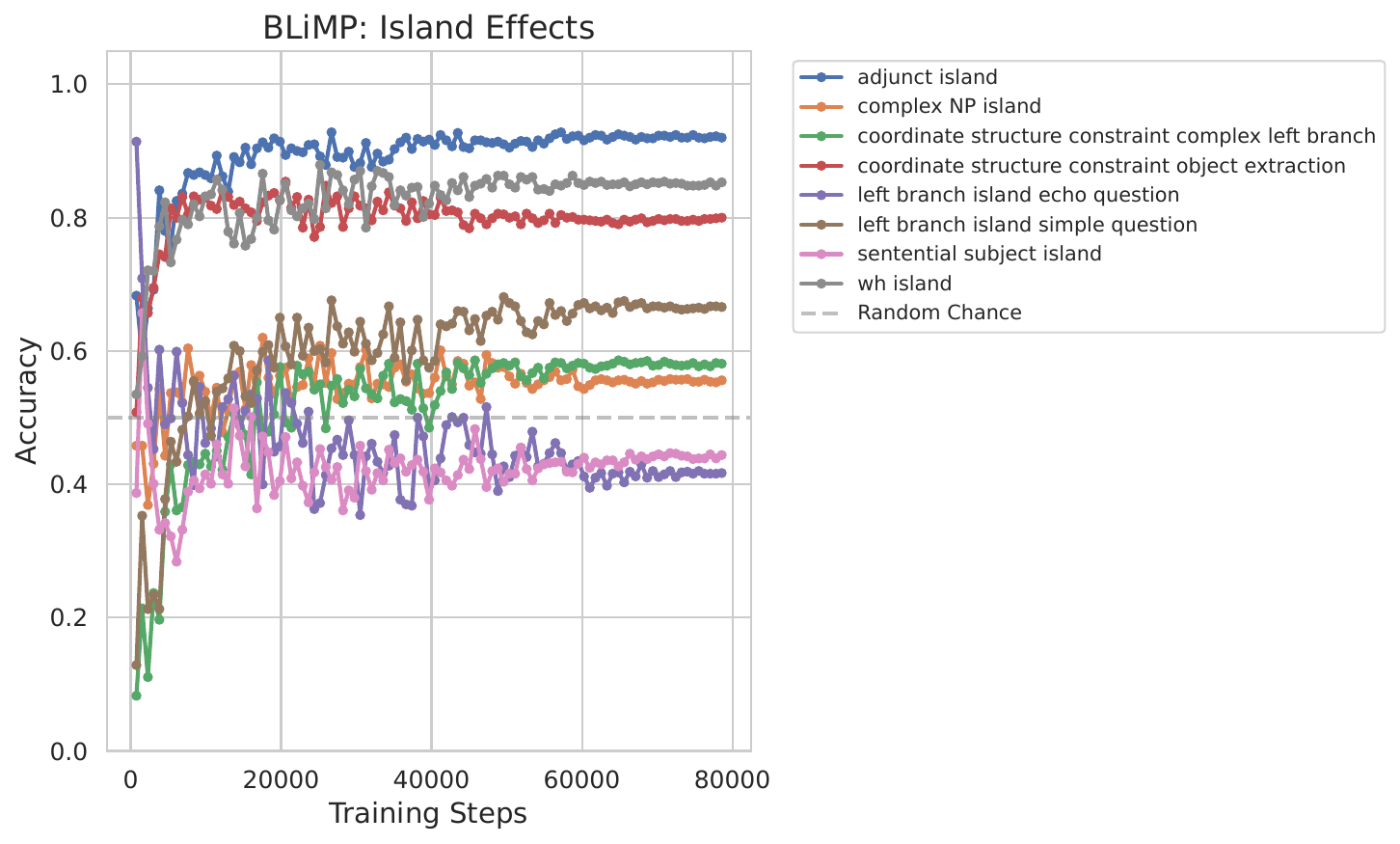}
    \caption*{Island Effects}
\end{minipage}

\vspace{0.4cm}

\begin{minipage}[t]{0.48\linewidth}
    \centering
    \includegraphics[width=\linewidth]{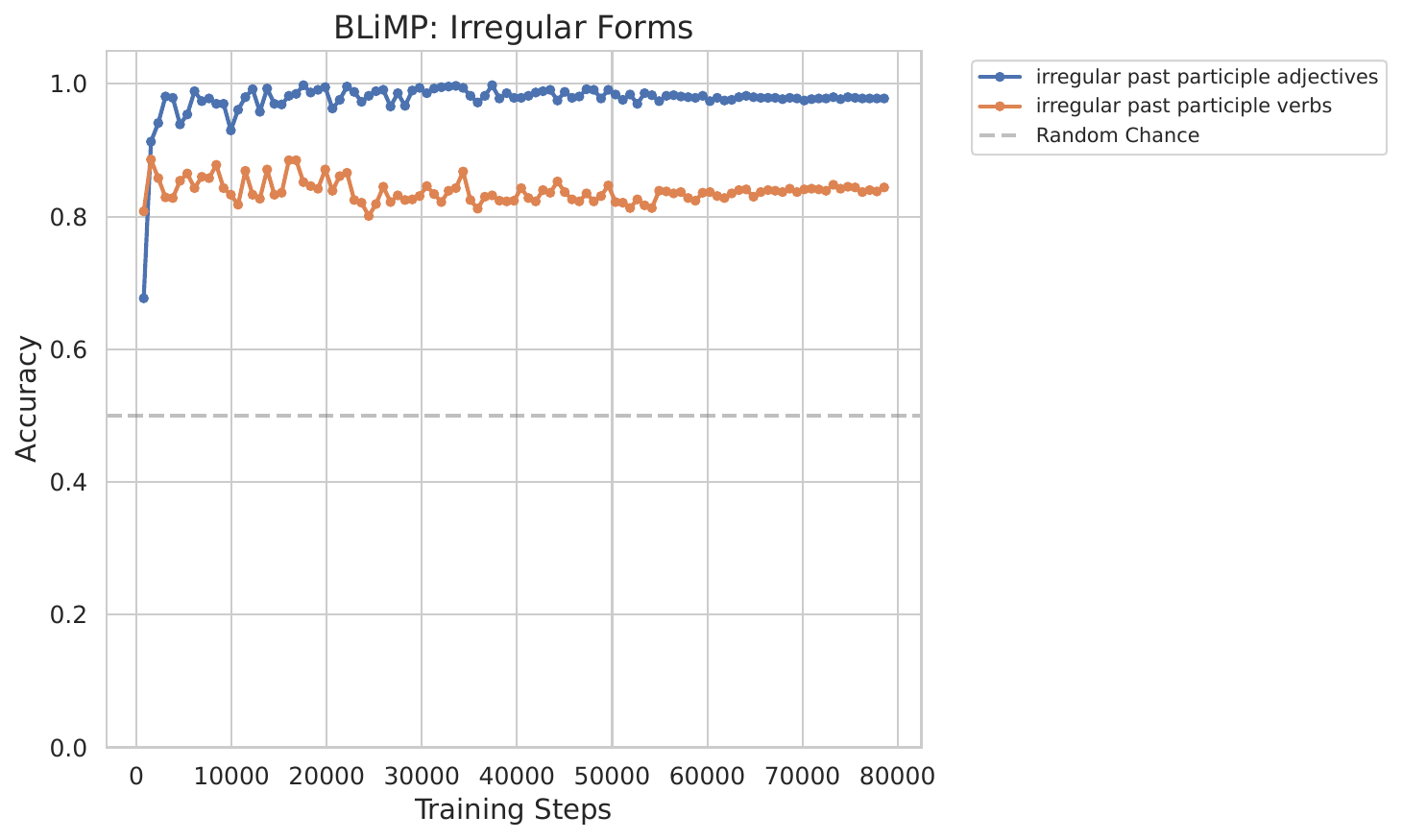}
    \caption*{Irregular Forms}
\end{minipage}\hfill
\begin{minipage}[t]{0.48\linewidth}
    \centering
    \includegraphics[width=\linewidth]{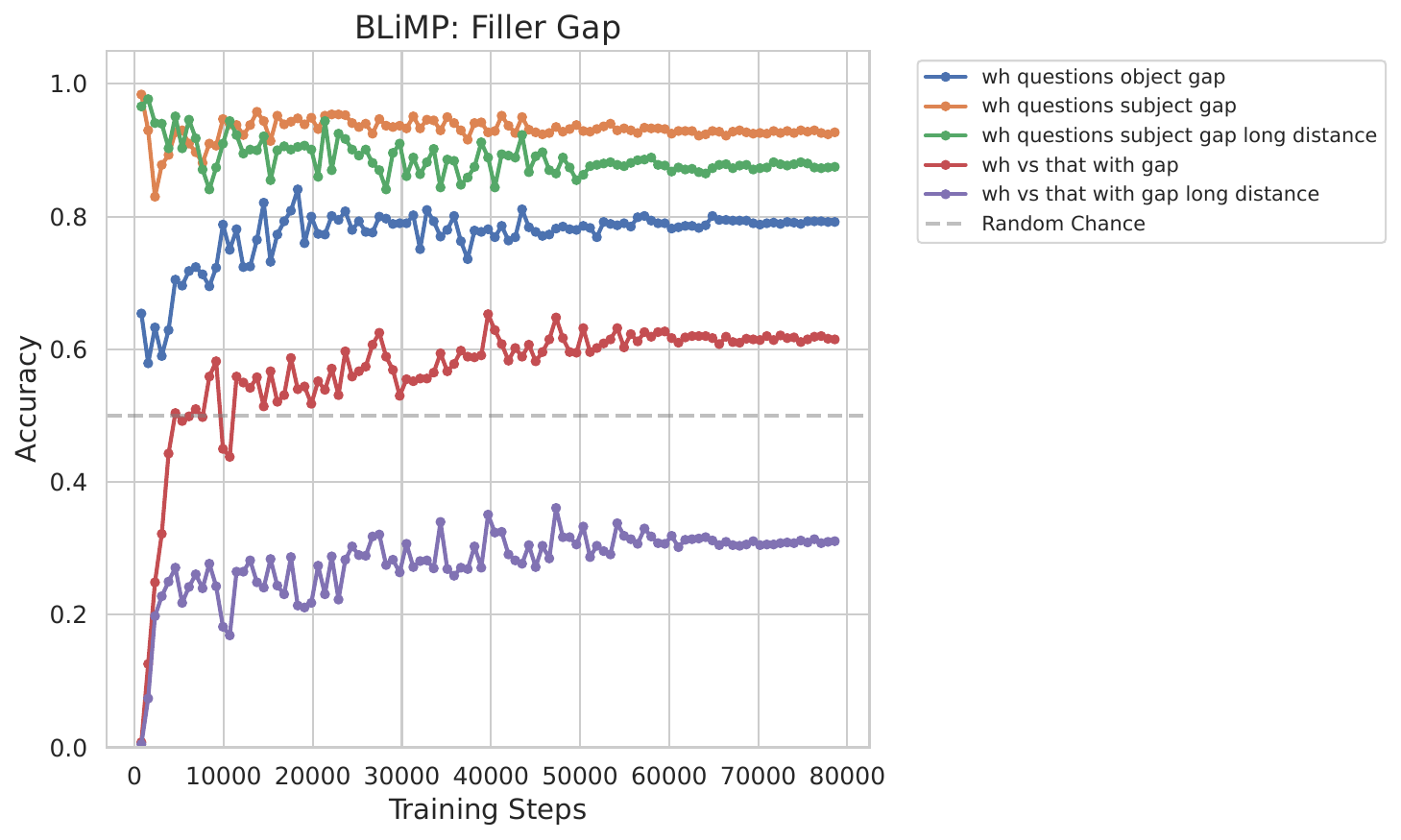}
    \caption*{Filler--Gap Dependencies}
\end{minipage}

\vspace{0.4cm}

\begin{minipage}[t]{0.48\linewidth}
    \centering
    \includegraphics[width=\linewidth]{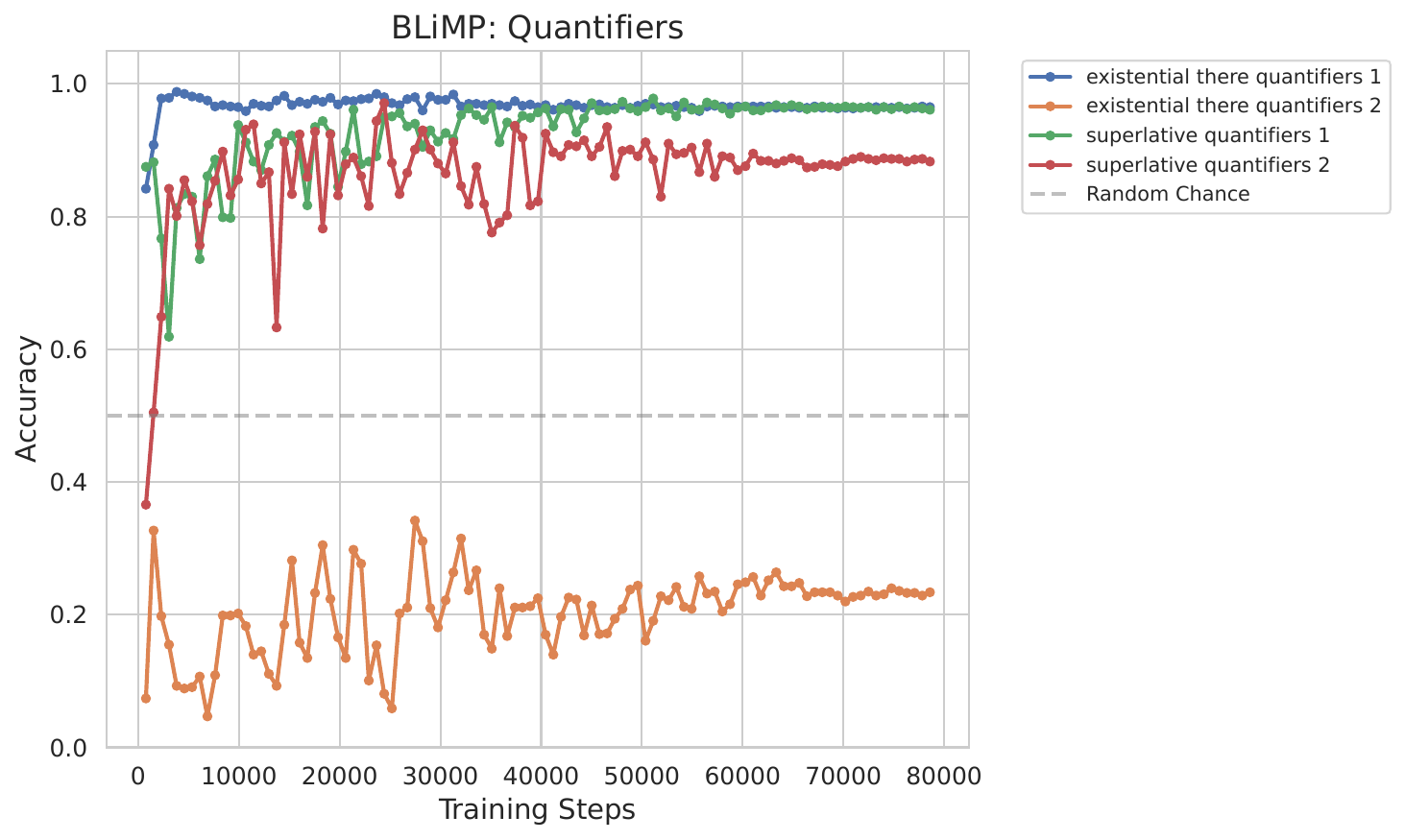}
    \caption*{Quantifiers}
\end{minipage}\hfill
\begin{minipage}[t]{0.48\linewidth}
    \centering
    \includegraphics[width=\linewidth]{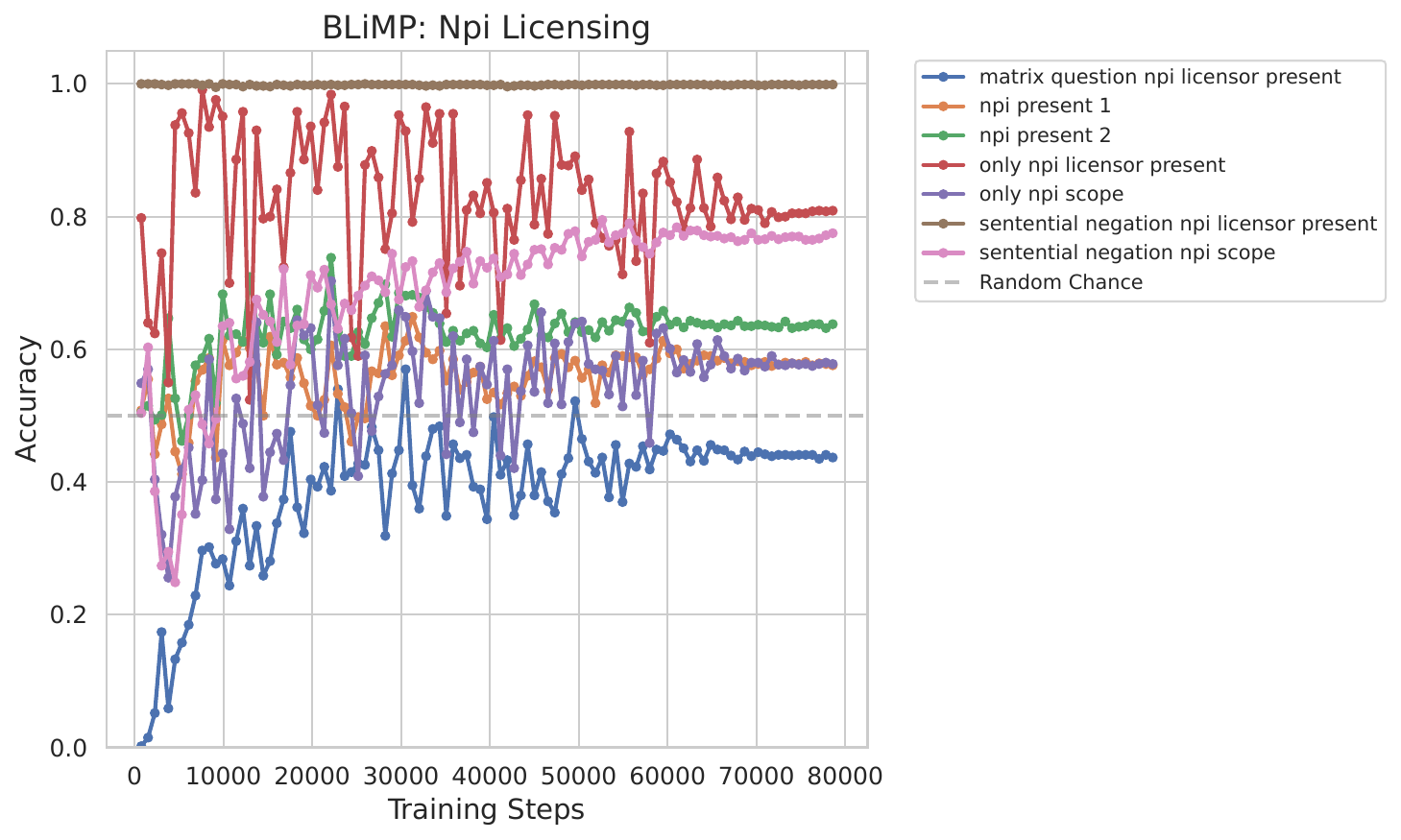}
    \caption*{NPI Licensing}
\end{minipage}

\caption{BLiMP results by linguistic phenomenon (1/2).}
\label{fig:blimp_1_2}
\end{figure}

\begin{figure}[H]
\centering

\begin{minipage}[t]{0.48\linewidth}
    \centering
    \includegraphics[width=\linewidth]{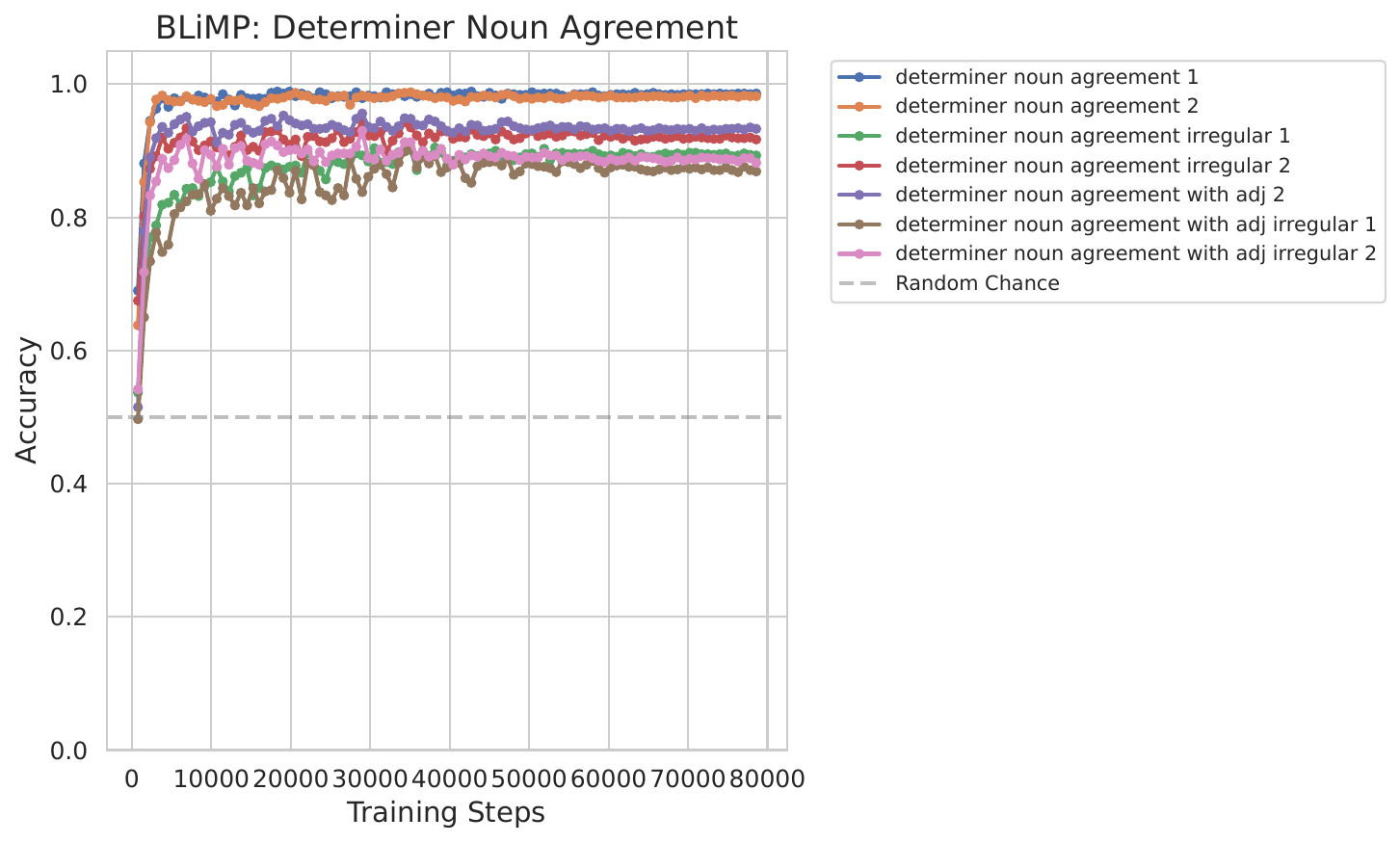}
    \caption*{Determiner--Noun Agreement}
\end{minipage}\hfill
\begin{minipage}[t]{0.48\linewidth}
    \centering
    \includegraphics[width=\linewidth]{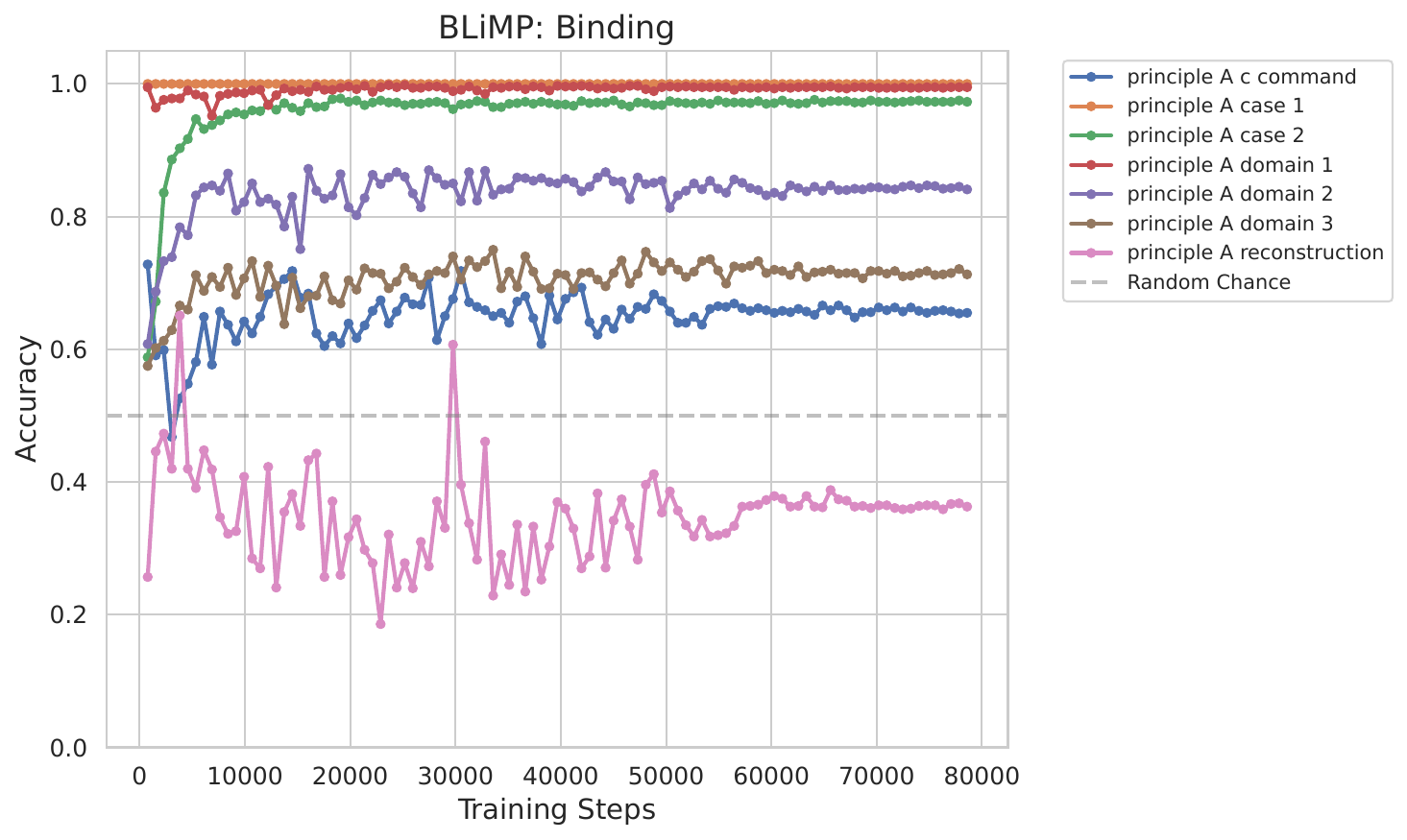}
    \caption*{Binding}
\end{minipage}

\vspace{0.4cm}

\begin{minipage}[t]{0.48\linewidth}
    \centering
    \includegraphics[width=\linewidth]{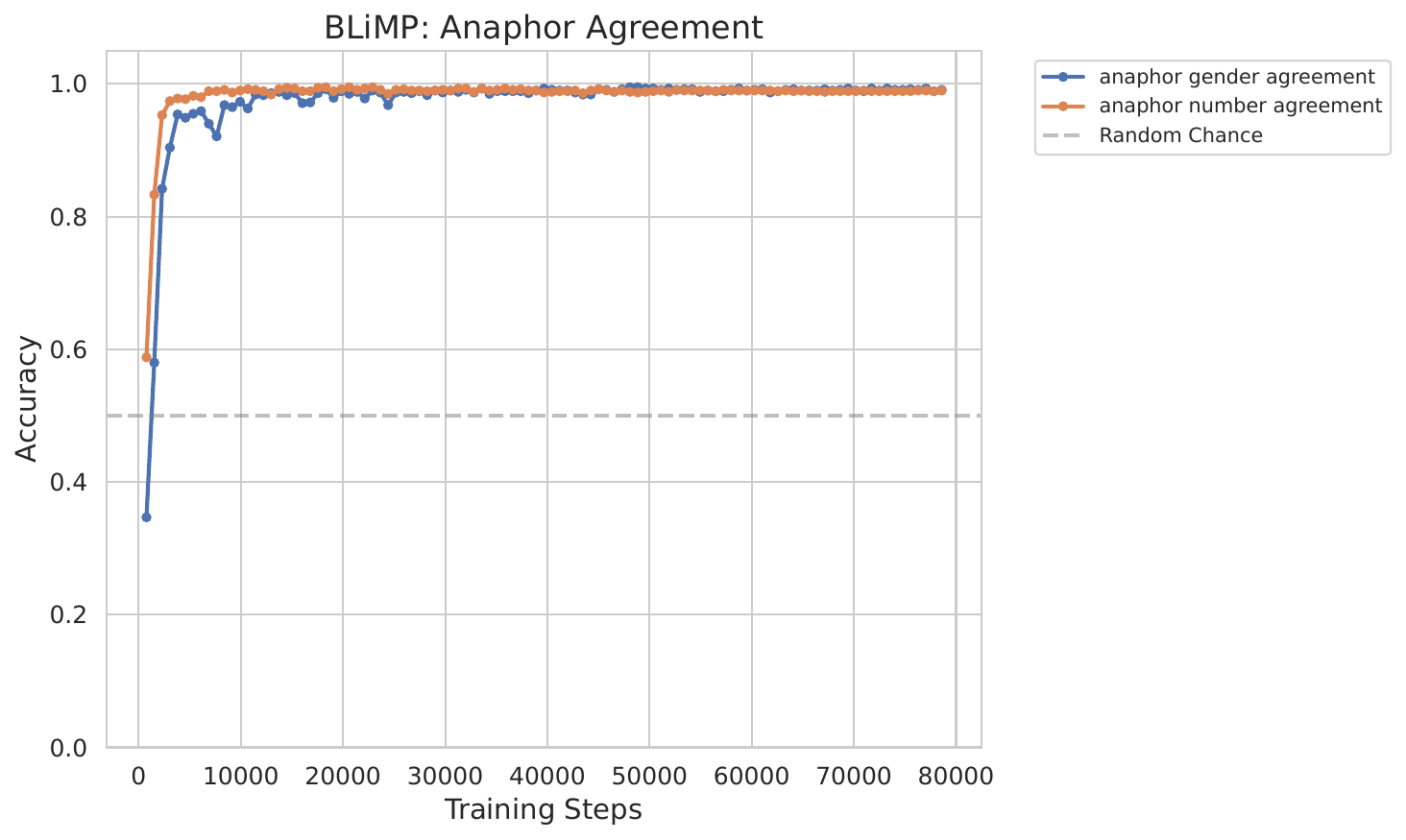}
    \caption*{Anaphor Agreement}
\end{minipage}\hfill
\begin{minipage}[t]{0.48\linewidth}
    \centering
    \includegraphics[width=\linewidth]{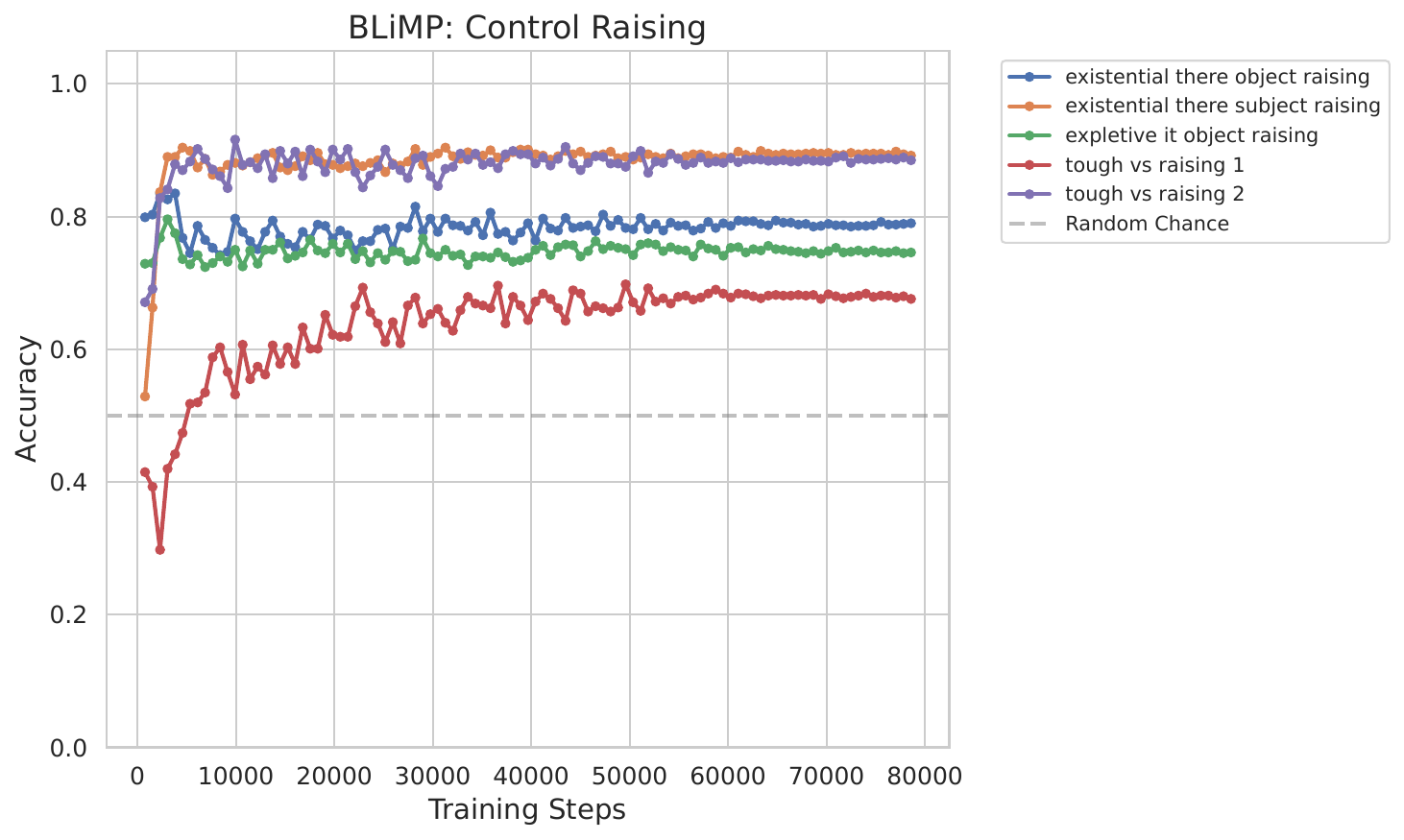}
    \caption*{Control and Raising}
\end{minipage}

\vspace{0.4cm}

\begin{minipage}[t]{0.48\linewidth}
    \centering
    \includegraphics[width=\linewidth]{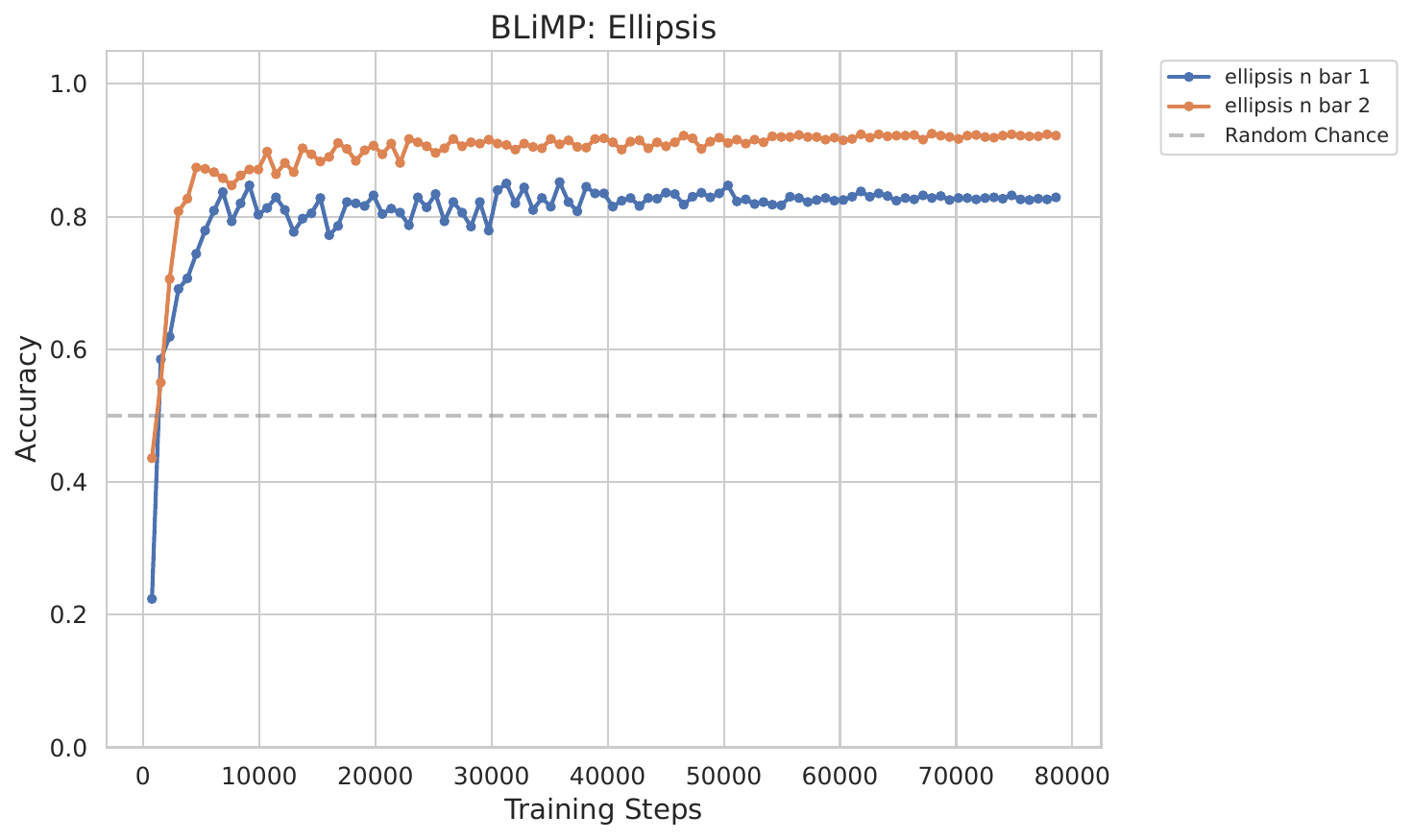}
    \caption*{Ellipsis}
\end{minipage}\hfill
\begin{minipage}[t]{0.48\linewidth}
    \centering
    \includegraphics[width=\linewidth]{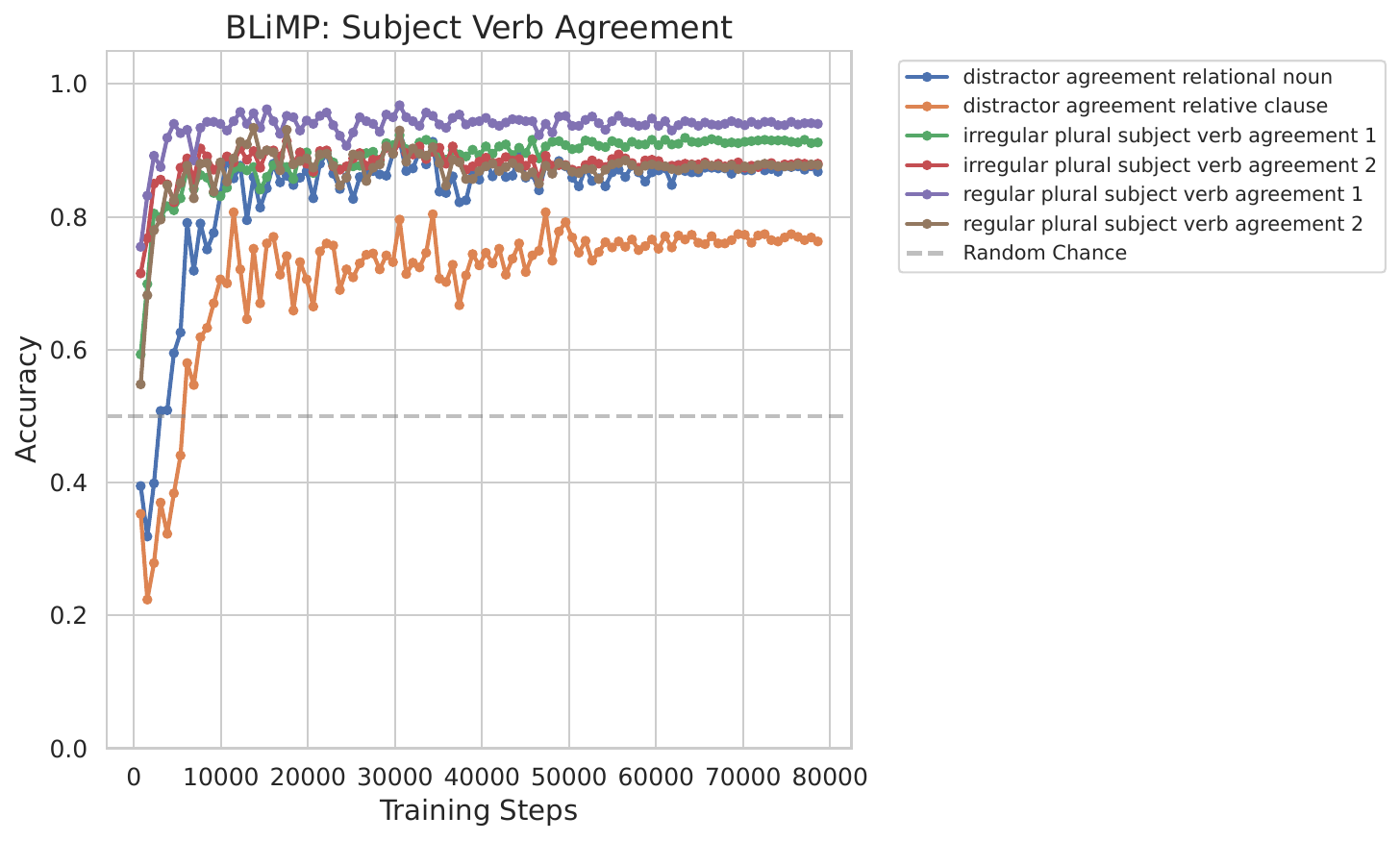}
    \caption*{Subject--Verb Agreement}
\end{minipage}

\caption{BLiMP evaluation results by linguistic phenomenon (2/2).}
\label{fig:blimp_2_2}
\end{figure}

Figures~\ref{fig:blimp_1_2} and ~\ref{fig:blimp_2_2} depict the learning curves of all the 67 BLiMP paradigms grouped by the 12 linguistic phenomena (when training a GPT-2 124M model on 10B tokens randomly sampled from FineWeb). As evident from these curves, some linguistic phenomena are captured very well, while some linguistic phenomena such as \texttt{existential\_there\_quantifiers\_2} and \texttt{principle\_A\_reconstruction} have significantly below chance accuracies -- with the latter also showing an inversed U-shaped scaling curve with performance increasing in early stages of training and decreasing gradually in the subsequent stages. 

\section{Synthetic Data Generation}
\label{app:synthetic_data_generation}
\subsection{Genres, Subgenres}
\label{app:synthetic_data_generation_genres}
Following is a taxonomy of genres we curate, for ensuring stylistic diversity in the generated synthetic data. Each genre also has a more fine-grained subgenre defined within itself. While generating each document in the synthetic data generation step, we sample one of these genres at random and also select a subgenre within it, at random. Please refer to our \href{https://github.com/kowndinya-renduchintala/heterogeneity-in-formal-linguistic-competence}{GitHub repository} for an extensive list of these subgenres considered.
\begin{itemize}[nosep]
\item Conversational and spoken
\item Narrative and storytelling
\item Descriptive (expository description)
\item Expository/informational
\item Instructional/procedural
\item Persuasive/argumentative
\item Academic/scholarly
\item Technical/scientific and engineering
\item Legal/regulatory
\item Business/professional
\item News/journalism
\item Government/public administration
\item Medical/healthcare
\item Finance/economics
\item Technology/consumer product
\item Culture/religion/philosophy
\item Sports
\item Entertainment/media
\item Social media/digital discourse
\item Education/pedagogy
\item Everyday documents/forms
\item Advertising/marketing
\item Customer support/service
\item Fictional micro-forms
\item Children's/young audience
\item Historical/archival styles
\item Humor/satire
\item Poetry and verse (prose-poem compatible)
\item Lists and structured text
\item Travel and geography
\item Environment/agriculture
\item Arts and literature
\item Real estate and housing
\item Transportation and logistics
\item Security/privacy/compliance
\item Gaming
\item Food and beverage
\item Fashion and lifestyle
\item Science communication (popular)
\end{itemize}

\subsection{Prompt used for Synthetic Data Generation}
\label{app:synthetic_data_generation_prompt}
Following is the prompt we use to generate the 1\% synthetic data using GPT-OSS 120B model:
\begin{wrapverbatim}
You are an expert linguist tasked with generating high-quality pre-training data for a small language model that specializes in formal linguistic competence (i.e., the knowledge of rules and statistical regularities of language).

You will be given:
- A linguistic paradigm along with its syntactic template.
- A target genre and a sub-genre.
- A set of lemmas that must appear in the text.

Your task is to generate a natural, fluent text that must include a sentence that demonstrates the given paradigm, and reads like a genuine piece of writing in the specified genre/sub-genre.

PARADIGM: {paradigm}
TEMPLATE: {template}

TARGET GENRE: {genre}
TARGET SUB-GENRE: {subgenre}

REQUIRED LEMMAS:
{formatted_lemmas}

INSTRUCTIONS:
- Do **not** mention grammar, linguistics, rules, or examples.
- Do **not** explain or comment on any construction.
- Avoid didactic tone, meta-language, or contrived sentence patterns.
- Preserve coherence, plausibility, and stylistic consistency appropriate to the genre.

Now go, generate the text.

OUTPUT:
\end{wrapverbatim}

\subsection{Syntactic Templates for BLiMP paradigms}
\label{app:synthetic_data_generation_templates}
Following are the syntactic templates that we use while generating synthetic data, for each of the 9 worst performing BLiMP paradigms that we consider in our experiments.

\begin{table}[h!]
\centering
\small
\begin{tabularx}{\linewidth}{l X}
\toprule
\textbf{Paradigm} & \textbf{Template} \\
\midrule

\texttt{only\_npi\_scope} &
Only \{subject\_NP\} \{relativizer\} \{subject\_embedded\}
\{auxiliary\_embedded\} \{verb\_embedded\} \{auxiliary\}
ever \{verb\_phrase\} \\
\hline
\texttt{existential\_there\_quantifiers\_2} &
\{all / every / each / most\} \{subject\_NP\} \{auxiliary\}
there \{verb\_phrase\} \\
\hline
\texttt{wh\_vs\_that\_with\_gap\_long\_distance} &
\{subject\_NP\} \{verb\} \{wh\_word\} \{subject\_embedded\}
that \{verb\_phrase\_relative\} \{verb\_embedded\} \\
\hline
\texttt{matrix\_question\_npi\_licensor\_present} &
\{auxiliary\} \{subject\_NP\} ever \{verb\_phrase\} \\
\hline
\texttt{principle\_A\_reconstruction} &
It's \{reflexive\} \{relativizer\} \{subject\_NP\}
\{verb\_phrase\} \\
\hline
\texttt{sentential\_subject\_island} &
\{wh\_word\} \{verb\_do\} \{\{subject\_NP\_possessive\}
\{verb\_ing\} \{subject\_NP\}\} \{verb / verb\_phrase\} \\
\hline
\texttt{left\_branch\_island\_echo\_question} &
\{subject\} \{auxiliary\} \{verb\}
\{wh\_determiner\} \{noun\} \\
\hline
\texttt{coordinate\_structure\_constraint\_complex\_left\_branch} &
\{wh\_determiner\} \{object\_noun\} \{auxiliary\}
\{subject\_1\} \{verb\_phrase\_1\} and
\{subject\_2\} \{verb\_phrase\_2\} \\
\hline
\texttt{principle\_A\_c\_command} &
\{subject\_NP\_1\} \{relativizer\} \{verb\_embedded\_2\}
\{subject\_NP\_embedded\_2\} \{verb\_phrase\_1\}
\{reflexive\} \\

\bottomrule
\end{tabularx}

\caption{Linguistic paradigms and their associated syntactic templates.}
\label{tab:paradigms-templates}
\end{table}


\section{Detailed Results}
\label{app:detailed_results}
\begin{table}[]
    \centering
    \resizebox{\textwidth}{!}{\begin{tabular}{lrrrrrrrrrrr}
\toprule
{} & \makecell{100\% \\FineWeb} &  \makecell{1\% synthetic\\random} &  \makecell{1\% synthetic\\coordinate\\structure\\constraint\\complex\_left\_branch} &  \makecell{1\% synthetic\\existential\\there\_quantifiers\_2} &  \makecell{1\% synthetic\\left\_branch\\island\_echo\_question} &  \makecell{1\% synthetic\\matrix\_question\\npi\_licensor\_present} &  \makecell{1\% synthetic\\only\_npi\_scope} &  \makecell{1\% synthetic\\principle\_A\\c\_command} & \makecell{1\% synthetic\\principle\_A\\reconstruction} &  \makecell{1\% synthetic\\sentential\_subject\\island} &  \makecell{1\% synthetic\\wh\_vs\_that\\with\_gap\\long\_distance} \\
PARADIGM                                               &             &                       &                                                                       &                                                 &                                                  &                                                      &                               &                                       &                                            &                                           &                                                    \\
\midrule
only\_npi\_scope                                     &           20.90 &                 17.00 &                                              44.10 &                                           45.40 &                                            30.70 &                                               26.1 &                         69.40 &                                 34.90 &                                      37.00 &                                     28.80 &                                              26.70 \\
existential\_there\_quantifiers\_2                    &           23.40 &                 12.60 &                                              13.00 &                                           51.80 &                                             6.50 &                                               12.4 &                         19.50 &                                 14.90 &                                      21.80 &                                     11.30 &                                               7.60 \\
wh\_vs\_that\_with\_gap\_long\_distance                  &           23.80 &                 24.60 &                                              23.30 &                                           27.80 &                                            27.00 &                                               25.0 &                         33.30 &                                 26.30 &                                      27.60 &                                     25.80 &                                              41.20 \\
matrix\_question\_npi\_licensor\_present               &           30.80 &                 25.20 &                                              31.60 &                                           20.90 &                                            19.20 &                                               44.2 &                         29.20 &                                 38.30 &                                      35.80 &                                     34.90 &                                              20.80 \\
principle\_A\_reconstruction                         &           37.20 &                 44.70 &                                              42.30 &                                           42.80 &                                            28.10 &                                               41.8 &                         39.50 &                                 50.80 &                                      78.90 &                                     42.70 &                                              35.90 \\
sentential\_subject\_island                          &           40.40 &                 31.30 &                                              44.20 &                                           36.10 &                                            41.80 &                                               49.5 &                         41.30 &                                 42.50 &                                      39.00 &                                     52.20 &                                              43.30 \\
left\_branch\_island\_echo\_question                   &           45.30 &                 44.60 &                                              39.60 &                                           45.70 &                                            63.20 &                                               36.2 &                         42.90 &                                 46.80 &                                      41.80 &                                     36.80 &                                              55.90 \\
coordinate\_structure\_constraint\_complex\_left\_br... &           46.50 &                 40.70 &                                              62.40 &                                           43.70 &                                            42.10 &                                               43.5 &                         43.80 &                                 52.50 &                                      56.20 &                                     45.10 &                                              46.10 \\
principle\_A\_c\_command                              &           49.00 &                 50.40 &                                              47.50 &                                           46.10 &                                            44.90 &                                               51.9 &                         46.90 &                                 46.00 &                                      49.90 &                                     48.70 &                                              49.30 \\
left\_branch\_island\_simple\_question                 &           51.30 &                 50.90 &                                              62.10 &                                           55.40 &                                            58.30 &                                               50.1 &                         49.80 &                                 62.40 &                                      59.90 &                                     53.10 &                                              57.20 \\
npi\_present\_1                                      &           51.90 &                 51.80 &                                              53.30 &                                           56.90 &                                            52.30 &                                               48.8 &                         47.40 &                                 50.30 &                                      56.00 &                                     43.00 &                                              47.00 \\
wh\_vs\_that\_with\_gap                                &           52.20 &                 51.20 &                                              53.70 &                                           53.90 &                                            55.60 &                                               56.2 &                         60.90 &                                 54.30 &                                      58.20 &                                     51.20 &                                              51.90 \\
only\_npi\_licensor\_present                          &           55.10 &                 66.40 &                                              92.70 &                                           96.50 &                                            97.10 &                                               74.0 &                         98.40 &                                 96.80 &                                      91.40 &                                     87.70 &                                              77.50 \\
complex\_NP\_island                                  &           55.60 &                 57.00 &                                              52.00 &                                           54.20 &                                            57.90 &                                               54.6 &                         54.90 &                                 57.30 &                                      56.60 &                                     51.90 &                                              56.00 \\
inchoative                                         &           56.90 &                 61.00 &                                              57.40 &                                           58.00 &                                            58.20 &                                               59.4 &                         60.00 &                                 60.70 &                                      58.70 &                                     57.90 &                                              60.20 \\
tough\_vs\_raising\_1                                 &           57.30 &                 62.50 &                                              59.90 &                                           59.70 &                                            59.20 &                                               66.6 &                         60.90 &                                 62.00 &                                      57.10 &                                     56.40 &                                              54.90 \\
npi\_present\_2                                      &           58.60 &                 56.20 &                                              57.20 &                                           58.10 &                                            55.50 &                                               53.7 &                         50.30 &                                 54.80 &                                      56.60 &                                     55.50 &                                              52.10 \\
distractor\_agreement\_relative\_clause               &           61.10 &                 56.10 &                                              55.90 &                                           59.50 &                                            56.00 &                                               53.5 &                         53.70 &                                 59.50 &                                      58.00 &                                     56.60 &                                              56.60 \\
sentential\_negation\_npi\_scope                      &           62.60 &                 46.90 &                                              48.00 &                                           59.40 &                                            52.00 &                                               56.1 &                         49.40 &                                 53.40 &                                      55.60 &                                     46.50 &                                              42.60 \\
principle\_A\_domain\_3                               &           63.40 &                 62.70 &                                              67.10 &                                           62.50 &                                            67.20 &                                               65.9 &                         61.50 &                                 65.20 &                                      68.70 &                                     66.70 &                                              65.80 \\
wh\_island                                          &           67.80 &                 76.40 &                                              81.60 &                                           86.80 &                                            77.60 &                                               86.6 &                         84.00 &                                 88.30 &                                      78.30 &                                     81.40 &                                              83.90 \\
causative                                          &           69.60 &                 70.60 &                                              71.20 &                                           70.40 &                                            71.00 &                                               70.7 &                         73.90 &                                 72.20 &                                      71.60 &                                     71.30 &                                              68.90 \\
coordinate\_structure\_constraint\_object\_extraction  &           71.70 &                 82.30 &                                              77.50 &                                           83.80 &                                            79.30 &                                               76.8 &                         79.80 &                                 79.00 &                                      78.50 &                                     76.40 &                                              80.20 \\
wh\_questions\_object\_gap                            &           73.20 &                 71.40 &                                              72.30 &                                           73.00 &                                            68.60 &                                               71.1 &                         70.20 &                                 71.20 &                                      74.70 &                                     69.90 &                                              73.90 \\
expletive\_it\_object\_raising                        &           74.00 &                 74.00 &                                              71.20 &                                           74.30 &                                            73.00 &                                               71.3 &                         73.70 &                                 73.90 &                                      71.60 &                                     71.50 &                                              72.20 \\
drop\_argument                                      &           74.20 &                 76.40 &                                              74.20 &                                           77.00 &                                            72.30 &                                               76.2 &                         76.30 &                                 75.90 &                                      75.20 &                                     71.40 &                                              77.30 \\
intransitive                                       &           74.20 &                 75.80 &                                              75.00 &                                           74.90 &                                            71.00 &                                               73.8 &                         74.20 &                                 73.90 &                                      70.90 &                                     73.00 &                                              75.20 \\
principle\_A\_domain\_2                               &           74.20 &                 73.00 &                                              79.00 &                                           75.00 &                                            75.10 &                                               75.3 &                         74.10 &                                 74.90 &                                      77.10 &                                     76.80 &                                              79.80 \\
animate\_subject\_passive                            &           77.00 &                 78.70 &                                              79.30 &                                           80.10 &                                            80.30 &                                               82.0 &                         76.50 &                                 79.50 &                                      80.10 &                                     80.70 &                                              80.20 \\
distractor\_agreement\_relational\_noun               &           77.50 &                 79.80 &                                              76.60 &                                           80.50 &                                            81.30 &                                               74.9 &                         77.70 &                                 78.00 &                                      80.80 &                                     77.00 &                                              75.20 \\
existential\_there\_object\_raising                   &           78.50 &                 71.00 &                                              70.40 &                                           76.90 &                                            72.10 &                                               77.8 &                         76.70 &                                 79.40 &                                      72.90 &                                     79.40 &                                              81.60 \\
adjunct\_island                                     &           79.50 &                 82.40 &                                              79.90 &                                           79.60 &                                            80.80 &                                               82.2 &                         81.40 &                                 82.90 &                                      86.40 &                                     85.80 &                                              84.10 \\
ellipsis\_n\_bar\_1                                   &           79.80 &                 79.40 &                                              77.10 &                                           80.70 &                                            76.90 &                                               76.9 &                         79.80 &                                 77.60 &                                      75.20 &                                     77.60 &                                              79.60 \\
superlative\_quantifiers\_1                          &           82.20 &                 72.50 &                                              85.30 &                                           79.40 &                                            80.20 &                                               80.4 &                         88.80 &                                 79.30 &                                      73.80 &                                     84.00 &                                              77.50 \\
determiner\_noun\_agreement\_irregular\_1              &           82.70 &                 83.00 &                                              82.60 &                                           83.60 &                                            82.70 &                                               82.7 &                         83.20 &                                 83.90 &                                      81.90 &                                     80.90 &                                              84.20 \\
determiner\_noun\_agreement\_with\_adj\_irregular\_1     &           82.80 &                 85.20 &                                              83.40 &                                           79.80 &                                            86.60 &                                               84.8 &                         81.70 &                                 81.00 &                                      81.90 &                                     85.40 &                                              80.60 \\
transitive                                         &           83.60 &                 82.80 &                                              84.20 &                                           84.00 &                                            83.20 &                                               84.7 &                         83.10 &                                 84.00 &                                      84.80 &                                     84.80 &                                              82.90 \\
regular\_plural\_subject\_verb\_agreement\_2            &           83.90 &                 87.50 &                                              86.40 &                                           84.60 &                                            85.90 &                                               88.0 &                         88.20 &                                 88.40 &                                      89.20 &                                     84.80 &                                              82.10 \\
irregular\_plural\_subject\_verb\_agreement\_2          &           84.40 &                 84.30 &                                              85.60 &                                           84.00 &                                            86.90 &                                               88.3 &                         82.60 &                                 84.30 &                                      86.30 &                                     85.10 &                                              82.60 \\
existential\_there\_subject\_raising                  &           84.80 &                 87.50 &                                              84.60 &                                           86.10 &                                            88.10 &                                               88.6 &                         86.30 &                                 86.90 &                                      86.10 &                                     89.60 &                                              88.00 \\
irregular\_plural\_subject\_verb\_agreement\_1          &           85.60 &                 87.50 &                                              86.50 &                                           86.90 &                                            88.50 &                                               86.9 &                         88.40 &                                 86.10 &                                      86.80 &                                     86.90 &                                              87.00 \\
determiner\_noun\_agreement\_with\_adj\_irregular\_2     &           85.90 &                 88.90 &                                              86.30 &                                           86.40 &                                            89.70 &                                               89.2 &                         87.20 &                                 87.80 &                                      85.50 &                                     87.50 &                                              83.50 \\
ellipsis\_n\_bar\_2                                   &           86.10 &                 86.40 &                                              84.30 &                                           83.50 &                                            83.50 &                                               84.0 &                         86.90 &                                 84.70 &                                      83.20 &                                     84.10 &                                              84.70 \\
irregular\_past\_participle\_verbs                    &           87.30 &                 84.00 &                                              83.50 &                                           83.10 &                                            84.30 &                                               82.4 &                         83.90 &                                 88.10 &                                      86.30 &                                     88.50 &                                              83.30 \\
tough\_vs\_raising\_2                                 &           87.90 &                 88.10 &                                              86.10 &                                           86.80 &                                            88.70 &                                               88.1 &                         86.90 &                                 87.60 &                                      89.00 &                                     88.80 &                                              91.30 \\
passive\_1                                          &           88.80 &                 88.90 &                                              90.60 &                                           86.80 &                                            89.20 &                                               87.9 &                         89.80 &                                 90.10 &                                      89.50 &                                     88.10 &                                              88.50 \\
determiner\_noun\_agreement\_irregular\_2              &           89.00 &                 92.00 &                                              88.60 &                                           89.70 &                                            92.90 &                                               90.0 &                         88.10 &                                 91.60 &                                      89.30 &                                     89.20 &                                              91.00 \\
determiner\_noun\_agreement\_with\_adj\_2               &           89.20 &                 91.40 &                                              89.10 &                                           89.80 &                                            93.00 &                                               92.0 &                         91.20 &                                 91.40 &                                      90.40 &                                     91.10 &                                              89.60 \\
passive\_2                                          &           90.20 &                 90.50 &                                              91.00 &                                           91.00 &                                            91.60 &                                               91.0 &                         90.80 &                                 91.10 &                                      89.60 &                                     90.10 &                                              91.30 \\
regular\_plural\_subject\_verb\_agreement\_1            &           90.40 &                 90.20 &                                              90.60 &                                           89.90 &                                            90.50 &                                               90.9 &                         89.90 &                                 91.00 &                                      92.00 &                                     90.60 &                                              92.40 \\
wh\_questions\_subject\_gap\_long\_distance             &           90.80 &                 88.90 &                                              93.80 &                                           87.90 &                                            89.50 &                                               88.5 &                         86.60 &                                 83.40 &                                      85.00 &                                     89.10 &                                              83.20 \\
wh\_questions\_subject\_gap                           &           91.20 &                 89.50 &                                              90.40 &                                           89.50 &                                            86.50 &                                               89.6 &                         86.60 &                                 86.20 &                                      88.30 &                                     88.30 &                                              87.40 \\
irregular\_past\_participle\_adjectives               &           91.40 &                 96.30 &                                              93.40 &                                           96.00 &                                            74.10 &                                               89.6 &                         97.30 &                                 96.90 &                                      97.80 &                                     98.40 &                                              93.10 \\
animate\_subject\_trans                              &           92.00 &                 90.90 &                                              90.50 &                                           91.20 &                                            90.50 &                                               88.8 &                         88.40 &                                 89.90 &                                      91.30 &                                     88.90 &                                              91.80 \\
determiner\_noun\_agreement\_with\_adjective\_1         &           93.90 &                 94.70 &                                              94.00 &                                           94.00 &                                            95.30 &                                               94.0 &                         93.70 &                                 94.10 &                                      93.20 &                                     93.60 &                                              92.90 \\
wh\_vs\_that\_no\_gap                                  &           95.00 &                 93.70 &                                              95.40 &                                           94.50 &                                            94.60 &                                               93.8 &                         93.40 &                                 94.80 &                                      95.30 &                                     95.00 &                                              94.00 \\
principle\_A\_case\_2                                 &           95.60 &                 94.80 &                                              91.90 &                                           95.10 &                                            95.10 &                                               94.6 &                         92.60 &                                 93.20 &                                      94.00 &                                     94.00 &                                              94.30 \\
superlative\_quantifiers\_2                          &           95.90 &                 92.40 &                                              83.70 &                                           91.20 &                                            91.10 &                                               82.5 &                         89.80 &                                 87.50 &                                      85.80 &                                     93.00 &                                              87.00 \\
anaphor\_gender\_agreement                           &           96.50 &                 94.00 &                                              97.20 &                                           96.00 &                                            97.10 &                                               97.1 &                         97.00 &                                 97.20 &                                      97.60 &                                     97.40 &                                              95.40 \\
determiner\_noun\_agreement\_2                        &           97.50 &                 97.10 &                                              95.70 &                                           96.60 &                                            96.80 &                                               97.0 &                         97.20 &                                 97.10 &                                      97.40 &                                     97.80 &                                              96.10 \\
wh\_vs\_that\_no\_gap\_long\_distance                    &           97.50 &                 95.40 &                                              96.30 &                                           97.20 &                                            95.80 &                                               96.7 &                         96.80 &                                 96.80 &                                      96.40 &                                     96.20 &                                              92.40 \\
existential\_there\_quantifiers\_1                    &           98.20 &                 97.80 &                                              96.20 &                                           97.70 &                                            96.60 &                                               97.8 &                         97.60 &                                 95.50 &                                      97.30 &                                     97.10 &                                              97.50 \\
determiner\_noun\_agreement\_1                        &           98.50 &                 98.00 &                                              98.10 &                                           98.40 &                                            98.00 &                                               98.0 &                         97.20 &                                 98.20 &                                      98.30 &                                     98.00 &                                              97.50 \\
principle\_A\_domain\_1                               &           98.50 &                 98.90 &                                              97.80 &                                           97.60 &                                            99.10 &                                               99.3 &                         98.90 &                                 97.10 &                                      98.50 &                                     98.20 &                                              98.50 \\
anaphor\_number\_agreement                           &           98.80 &                 98.60 &                                              99.20 &                                           98.80 &                                            99.00 &                                               99.4 &                         99.00 &                                 99.00 &                                      98.70 &                                     99.30 &                                              98.80 \\
sentential\_negation\_npi\_licensor\_present           &           99.00 &                 99.60 &                                              98.50 &                                           99.60 &                                            99.60 &                                              100.0 &                         99.30 &                                100.00 &                                      99.90 &                                     99.90 &                                              98.70 \\
principle\_A\_case\_1                                 &          100.00 &                100.00 &                                             100.00 &                                          100.00 &                                           100.00 &                                              100.0 &                        100.00 &                                100.00 &                                     100.00 &                                    100.00 &                                             100.00 \\
\hline
\textbf{BLiMP (Overall Score)}                                              &           \textbf{74.77} &                 \textbf{74.57} &                                              \textbf{75.72} &                                           \textbf{76.48} &                                            \textbf{75.17} &                                               \textbf{75.4} &                         \textbf{76.11} &                                 \textbf{76.43} &                                      \textbf{76.69} &                                     \textbf{75.35} &                                              \textbf{74.92} \\
\bottomrule
\end{tabular}

}
    \caption{Detailed breakdown of scores on 67 individual  BLiMP paradigms}
    \label{tab:detailed_results}
\end{table}

Table~\ref{tab:detailed_results} contains detailed scores for all the 67 BLiMP paradigms, corresponding to our data intervention experiments described in Section~\ref{sec:results_n_discussion}.

\section{Use of AI Assistants}

The authors acknowledge the use of AI assistants for editing some parts of the paper and declare that none of the generated content is presented without rigorous manual review.

\end{document}